\documentclass[lettersize,journal]{IEEEtran}

\usepackage{amsmath,amsfonts}
\usepackage{algorithmic}
\usepackage{algorithm}
\usepackage{array}
\usepackage[caption=false,font=normalsize,labelfont=sf,textfont=sf]{subfig}
\usepackage{textcomp}
\usepackage{stfloats}
\usepackage{url}
\usepackage{verbatim}
\usepackage{graphicx}
\usepackage{cite}
\usepackage{tabularx} 
\usepackage{units}
\usepackage{xcolor}

\usepackage{tikz} 
\usetikzlibrary{arrows,automata}

\begin{document}
%
\title{Minimalist exploration strategies for robot swarms at the edge of chaos}
%
%
%


\author{Vinicius Sartorio$^{1}$, Luigi Feola$^{2}$, Emanuel Estrada$^{3}$, Vito Trianni$^{2}$ and Jonata Tyska Carvalho$^{1}$
\thanks{$^{1}$Vinicius Sartorio and Jonata Tyska Carvalho is with Department of Informatics and Statistics, Federal University of Santa Catarina, R. Eng. Agronômico Andrei Cristian Ferreira, Brazil {\tt\small vinicius.sartorio@posgrad.ufsc.br}}%
\thanks{$^{2}$Luigi Feola and Vito Trianni with the Institute of Science and Technology of Cognition, National Research Council, Piazzale Aldo Moro, 7 - 00185 Roma, Italia}%
\thanks{$^{3}$Emanuel Estrada is with Federal University of Rio Grande, Km 8 Avenida Itália Carreiros, Brazil}%
}


%
%

\markboth{IEEE ROBOTICS AND AUTOMATION LETTERS, VOL. , NO. , MONTH 2024}%
{Shell \MakeLowercase{\textit{et al.}}: Bare Demo of IEEEtran.cls for IEEE Journals}
%



\maketitle
%
\begin{abstract}
Effective exploration abilities are fundamental for robot swarms, especially when small, inexpensive robots are employed (e.g., micro- or nano-robots). Random walks are often the only viable choice if robots are too constrained regarding sensors and computation to implement state-of-the-art solutions. However, identifying the best random walk parameterisation may not be trivial. Additionally, variability among robots in terms of motion abilities---a very common condition when precise calibration is not possible---introduces the need for flexible solutions. This study explores how random walks that present chaotic or edge-of-chaos dynamics can be generated. We also evaluate their effectiveness for a simple exploration task performed by a swarm of simulated Kilobots. First, we show how Random Boolean Networks can be used as controllers for the Kilobots, achieving a significant performance improvement compared to the best parameterisation of a L\'evy-modulated Correlated Random Walk. Second, we demonstrate how chaotic dynamics are beneficial to maximise exploration effectiveness. Finally, we demonstrate how the exploration behavior produced by Boolean Networks can be optimized through an Evolutionary Robotics approach while maintaining the chaotic dynamics of the networks.
\end{abstract}



\section{Introduction}\label{introduction}

Robot swarms are expected to revolutionise many important applications~\cite{garate:2021,polvara:21}, owing to efficient parallel operation, robustness and scalability~\cite{10.1109/jproc.2021.3072740}. Futuristic visions prospect large-scale swarms and extreme miniaturisation, down from centimeter to nano-scale~\cite{10.1126/scirobotics.abe4385,7067029}. It is challenging to integrate complex sensing or actuation at similar sizes, as well as large computational abilities~\cite{10.1038/s41578-018-0001-3,10.1002/adma.201703554}. To preserve the robot's autonomy, it is therefore of fundamental importance to abandon tight control schemes in favour of minimalist approaches~\cite{Gauci:2014kb,8264725,10.1109/lra.2022.3150479}.

Coordinated navigation and exploration are critical abilities for robot swarms~\cite{10.1126/scirobotics.aaw9710,10.1126/scirobotics.abd8668}. Minimalist approaches do not rely on maps and self-localisation~\cite{10.3389/frobt.2021.618268}, but rather exploit stochastic search strategies based on Random Walks \cite{bartumeus2005animal,Codling2008,Levy2015}. 
RWs can be implemented with minimal complexity, exploiting either external stochastic fluctuations~\cite{Angelani_2010} or simple controllers that only require a random number generator to produce the stochastic distributions that determine the sequence of step lengths and turning angles. By controlling these two distributions, a wide range of RW patterns can be obtained, from simple Brownian motion to correlated RW (CRW) and Lévy walks (LW), offering the possibility to choose the best search pattern for the task at hand. For instance, a mild CRW proves best for target search in case of frequent collisions with boundaries, while a LW should be preferred for search problems in open space~\cite{Dimidov2016}. 
Several studies utilizing RWs are found in the literature. To name a few, in~\cite{Hecker2015}, a biased CRW was implemented to solve a collective foraging task; in~\cite{Kegeleirs2019}, RWs were used as mapping strategies in a close-space scenario; and in ~\cite{Reina2015}, a CRW was used by robots in a collective decision-making problem. 

Generally speaking, the best parameterisation of a RW for exploration with robot swarms largely depends on the problem characteristics, the robot embodiment, and the resulting motion ability. It is plain to admit that when choosing the parameterisation of the RW for a given task, there is no one-size-fits-all solution. Environmental dynamics and variability in the conditions leading to task execution may require radically different search patterns, which may not be predictable beforehand. Similarly, heterogeneity among the swarm robots may require flexible approaches to exploration. Intrinsic heterogeneity among robots is a critical, often overlooked aspect of swarm robotics. Practitioners are aware of the variability that exists among the robots composing a swarm, which can be reduced by careful calibration and control laws that correct individual errors. However, when moving towards extreme miniaturisation, calibration and control may not help. In a recent study~\cite{10.48550/arxiv.2305.16063}, experiments with Kilobots~\cite{kilobots} demonstrated that every single robot features consistent deviations from the expected behaviour, be it in the clock frequency, sensing, or motion. In particular, Kilobots have a different, time-varying heading bias during straight motion that leads robots to trace curved paths of varying radius. While this behaviour was known beforehand (e.g., a systematic heading bias was introduced in Kilobots' simulations~\cite{Pinciroli:2018}), the novel insight consists in the possibility to exploit such variations for better swarm behaviour: individual variations should be considered a feature rather than a bug~\cite{10.48550/arxiv.2305.16063}.

Bio-inspiration can be a powerful tool to exploit the full range of individual variations while maintaining a minimalist approach to behaviour control. For instance, many natural systems present chaotic or edge-of-chaos dynamics that lead to efficient behavioural patterns. \textit{Chaotic} dynamics are characterized by an aperiodic bounded behaviour extremely sensitive to initial conditions~\cite{kaplan1997}. \textit{edge-of-chaos} dynamics are found ``at the edge'' between order and chaos. Such dynamics can propagate small changes in the system when they are ``perturbed'' but at the same time preserve information~\cite{CarlosRBN}, meaning that changes are not completely random. As Langton proposed in~\cite{LANGTON1990}, living systems can find optimal conditions for life in a regime close to the edge of chaos since they get the required stability to survive and, simultaneously, can more effectively explore new possibilities. 
Indeed, these kinds of dynamics can be observed at different organizational levels, from intra-cellular formation and stabilization of microtubules, going through plant morphology, root and stem growth, to animal behavior such as the exploratory movements of \emph{Trogoderma variabile} beetles in their search for pheromones~\cite{west:03} (p.37-39). Many recent studies detected dynamics at the edge of chaos in different natural systems~\cite{morales:21,clark:20,zhang:20,ruan:19}. 

Exploiting chaotic or edge-of-chaos dynamics for robot control has not received sufficient attention to date. An interesting study implemented a Chaotic Random Bit Generator on a microcontroller of an autonomous mobile robot to enable coverage of the workspace within the fastest time possible~\cite{VOLOS2013}. In a later study, evolved robot's controllers producing chaotic dynamics lead to good performance also when placed in new, never-experienced environments~\cite{DaRold2015}. Here, we propose a minimalist approach to produce RW based on chaotic or edge-of-chaos dynamics, exploiting Random Boolean Networks (RBNs) as controllers. We perform experiments exploiting a realistic simulation of Kilobots (including individual variability), in which the swarm has to explore an unknown area and find a target object. Our results demonstrate that dynamics at the edge of chaos can significantly improve the exploration performance. We also show how performance can be maximized by an evolutionary optimisation process that finds the best possible controller while preserving relevant chaotic dynamics.

The remainder of this article is organized as follows.
Section~\ref{experimental_setup} presents the details of the tools and the experimental setup used for conducting the proposed investigation. We report and discuss the results obtained in Section~\ref{Results}, and we conclude the paper in Section~\ref{Conclusions} indicating possible future research directions.



\section{Experimental setup} \label{experimental_setup}
Our study aims to determine how a robot swarm can explore the greatest space possible in the fastest time. To this end, we devise an experiment in which robots must find a static target, and we measure the time taken by different individuals in the swarm to complete this task.

    \subsection{Experiment}
    \label{sec:experiment}

    Experiments are performed in simulation with the Kilobot robotic platform~\cite{kilobots}, a common choice in swarm robotics thanks to its low cost, small size, and ease of use. The simulation was implemented in the ARGoS simulator due to its high flexibility, efficiency, and fidelity in replicating real-life setups~\cite{Pinciroli:2012dc, Pinciroli:2018}. To account for the individual viability of Kilobots in their heading bias, a random velocity bias is introduced in the differential steering actuator. We use default values calibrated to physical Kilobots~\cite{Pinciroli:2018}, adding a random value drawn from a Gaussian distribution with average \unit[$\mu_v=0.00015$]{m/s} and standard deviation \unit[$\sigma_v=0.00270$]{m/s}.

    
    \begin{figure*}[t]
        \centering
        \includegraphics[width=\textwidth]{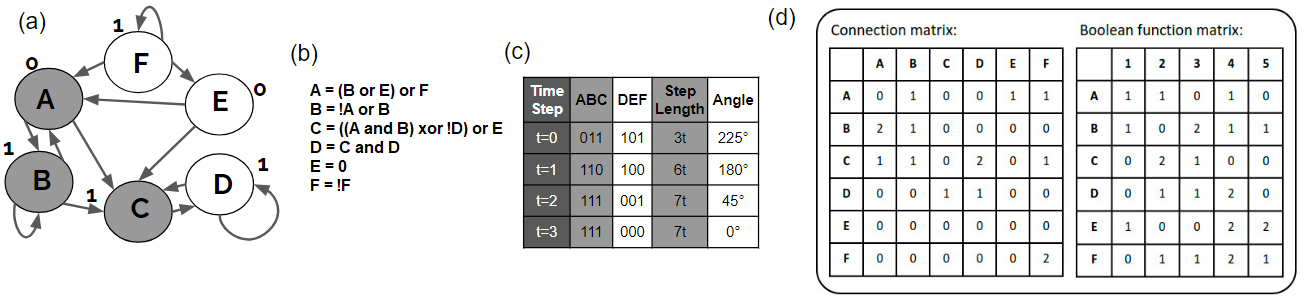}
        \caption{Representation of a 6 Node RBN. a) Graphic representation of a RBN showing connections and node states with "011101" initial state values. b) Boolean functions for each node of the network. c) Table showing node state values, robot step length value (in simulator ticks), and robot angle value (in degrees) for each time step (t=0 to t=3). d) RBN representation as an individual of the Genetic Algorithm population.}
        \label{fig:BNrepresentation}
    \end{figure*}

   A swarm of $R=20$ Kilobots searches for a single circular static target with radius $r_t=\unit[1.5]{cm}$ placed randomly in a circular bounded arena of radius $r_a=\unit[45]{cm}$. At the start, robots and targets are uniformly distributed in the arena. Kilobots do not have sensors to perceive obstacles and are, therefore, free to collide with each other and the walls. Robots are limited in their movement primitives: (i) straight motion, (ii) turning anticlockwise, (iii) turning clockwise, and (iv) stop motion. We consider that a robot finds the target when it approaches it at a distance closer than \unit[3]{cm}. Finding the target does not alter the robot's motion, which keeps moving according to the defined search strategy.

    The efficiency of the swarm exploration is measured as the average first passage time $t_f$, i.e., the average time for a robot to find the target for the first time. We evaluate $t_f$ for a given search strategy across $S=100$ simulation trials, changing the target's and robots' initial position every trial. Each trial lasts $T=\unit[3\times10^3]{s}$ to concede enough time for a good portion of the agents in the swarm to find the target and, therefore, compute a good approximation of $t_f$. To this end, we record the individual first passage times $t_f(r,s)$ of each robot $r$ in run $s$, obtaining at most $RS = 2000$ data points. Robots that do not encounter the target result in censored data points. Then, we estimate the cumulative distribution function (CDF) of the first passage times by using the Kaplan-Meier (K-M) estimator~\cite{kaplan1958nonparametric}. By examining our results in section~\ref{Results}, we concluded that the empirical CDF best fits a Weibull distribution for all the performed simulations. Hence, after fitting the Weibull distribution as shown in \eqref{eq:weibull} to the empirical CDF, we obtain the parameters $\alpha$ and $\gamma$ that can be used to compute the average first passage time $t_f$ as the mean of the fitted Weibull distribution, shown in \eqref{eq:meanw}.

    \begin{equation}\label{eq:weibull}
    F(x) = 1 - e^{-(\frac{x}{\lambda})^k},\quad x\geq 0
    \end{equation}

    \begin{equation}\label{eq:meanw}
        t_f = \lambda\Gamma(1 + \frac{1}{k})
    \end{equation}

    \subsection{Baseline search strategy}
    As a reference for a minimalist search strategy, we employ the Lévy-Modulated Correlated Random Walk (LMCRW), previously implemented for Kilobots~\cite{Dimidov2016}. The LMCRW is characterised by two parameters---$\rho$ and $\alpha$---respectively controlling the distribution of turning angles and step lengths. The turning angle is drawn from a wrapped Cauchy distribution with the following probability density function:
    \begin{equation}\label{eq:wcd}
        f(\theta,\rho) = \frac{1}{2\pi}\frac{1-\rho^2}{1+\rho^2+2\rho\cos(\theta)}
    \end{equation}
    The parameter $\rho$ determines the distribution width. The distribution becomes uniform when $\rho = 0$, and the RW is isotropic. When $\rho = 1$, a Dirac distribution is obtained, corresponding to ballistic motion.
    The step length $\delta$ follows a L\'{e}vy distribution characterised by a power law $P(\delta)\approx \delta^{-(\alpha+1)}$, with $0<\alpha\leq2$. For $\alpha = 2$, the distribution becomes Gaussian, while for $\alpha\rightarrow 0$, the RW reduces to ballistic motion.

    At every decision point, the robot draws a random turning angle $\theta$ and step length $\delta$. The robot then rotates at the given angle (with a rotation speed of approximately \unit[$\frac{\pi}{4}$]{s$^{-1}$}). Then, it moves straight to cover the desired distance (with a linear speed of approximately \unit[1]{cm/s}. Once the straight motion is completed, the robot chooses a new turning angle and step length.
    
    \subsection{Boolean Networks as robot behaviour control}
    Boolean Networks (BNs) have been introduced to model gene regulatory processes~\cite{KAUFFMAN1969}. A BN is a network with $N$ nodes characterised by a Boolean state. Each node is linked to the other nodes through $K$ directed edges, indicating that the receiver node is influenced by the state of $K$ source nodes. Edges also have a Boolean state, which indicates whether or not the state of the source node should be negated. Each node is characterised by a distinctive logic function, which is used to compute the node's new state from the input nodes' state. A BN is a discrete-state and discrete-time autonomous system, i.e., it evolves from its initial state without any external input~\cite{Shmulevich2009}. The network activation is synchronous: the states of nodes at time $t+1$ depend on the states of nodes at time $t$ according to the nodes' logic functions.
    
    A Random Boolean Network (RBN) is a BN with a random initial state, random connections, and randomly generated logic functions (see Figure~\ref{fig:BNrepresentation}a). Given the $K$ connections, $K-1$ logic gates (e.g., \textit{AND}, \textit{OR}, \textit{XOR}, etc) are randomly assigned to the node's logic function. For instance, when $K=3$, the logic function will have two Boolean operators in cascade, meaning that the first logic gate uses the first two inputs, and the second logic gate combines this result and the third input, as can be seen in Figure~\ref{fig:BNrepresentation}b). 
    Once the RBN is defined, all possible transient states of the given network can be computed. Since the state space is finite ($2^N$), eventually, a state could be reached again~\cite{CarlosRBN}. Many variations of the RBN models have been proposed over time. This study focuses on the non-homogeneous RBNs introduced by Derrida and Pomeau~\cite{Derrida1986}. Unlike the classical model proposed by Kauffman, each node can have a different number of connections, e.g., we can have a node with $K=5$ and another with $K=0$. Removing this constraint allows the exploration of a vast number of different networks and dynamics.

    In our study, a BN is represented by two matrices, one indicating connections and the other for logic functions (see Figure \ref{fig:BNrepresentation}d) for an example of a BN with $N=6$). The connections matrix is a $N\times N$ adjacency matrix, with each element of the connection matrix $c_{ij}$ denoting the type of connection between nodes $i$ and $j$: 0 when there is no connection, 1 when there is a simple connection, and 2 when there is a connection applying a boolean $NOT$. The matrix for the logical functions has the cardinality of $N\times(N-1)$, and the generic value $l_{ij}$ depends on the number of chosen logical gates. In this study, we only use \textit{AND}, \textit{OR}, and \textit{XOR} gates to reduce the search space complexity. Each value on this matrix represents a different Boolean function. The value zero represents an \textit{AND} function, one represents an \textit{OR} function, and two a \textit{XOR} function. Note that, for each node, only the first $K-1$ gates are used, where $K$ is the number of connections resulting from the connection matrix.

    In robotics, thanks to their compactness and readability, BNs have been used as control programs to perform tasks such as phototaxis, obstacle avoidance and even to generate finite state machines~\cite{Roli2011,Garattoni2013,Roli2013}. In these studies, BNs are synthesized through an automatic design methodology that uses an optimisation algorithm to modify the BN structure to fulfill a given task. In our study, a RBN controls the robot exploration behaviour. Each Kilobot in the experiment was assigned the same network configuration (same size, connections, and boolean functions). The only difference for each robot lies in the network's initial values. Each network controls the turning angle and the number of time-steps the robot moves straight. The output for these two parameters is extracted from the node states of the BN. As shown in Figure \ref{fig:BNrepresentation}(a), the BN is divided into two parts: one responsible for the step length $\delta$ and the other for the turning angle $\theta$. 
    To compute the time steps for straight motion, a simple conversion of binary to decimal is performed. Hence, the maximum duration is constrained between 0 and $2^{\frac{N}{2}}$ control ticks, where $N$ is the size of the BN. Instead, to compute the turning angle value, the binary network value is scaled in the range $[-\pi, \pi]$. After the new motion values are extracted, the network nodes are updated. In Figure~\ref{fig:BNrepresentation}c, we show an example of the step length duration and turning angle values for each time step for a network of size $N=6$.

    \subsection{Evolved Boolean Network}

    The RBNs can present highly variable results depending on the random connections and functions. To reduce this variability and achieve effective results regardless of the random initialization, we can optimise the network configuration through a Genetic Algorithm~\cite{mitchell1998}. Although combining Evolutionary Algorithms with RBN's has already been tried in the literature \cite{Ching2008AGA, roli2011boolean}, to the best of our knowledge, this is the first time this combination is investigated in the context of exploratory tasks. We call the resulting networks Evolved Boolean Networks (EBNs). 
    
    The evolutionary process operates on a population of $P = 40$ BNs with fixed node size $N$, which are initially generated randomly. 
    Each BN in the population is evaluated for the ability to minimise the average first passage time $t_f$, estimated with the same method described in Section~\ref{sec:experiment} over $S_e$ simulations trials. We use a small number of trials to speed up the evolutionary process, setting $S_e=8$. To ensure consistent evaluation conditions for all BNs in the population, we used the same set of $S_e$ random target positions for each individual. The robot's initial states are the same for each robot across all individuals and generations.
    
    After the evaluation process, a new population is generated from the old one. Selection is performed using a binary tournament for the population, in which two individuals participate in each competition comparing their fitness, and the one with lower $t_f$ is chosen to generate the following population. A simulated binary crossover operator(SBX)~\cite{GADeb2007} combines the genes of these parents to create one offspring following an exponential distribution probability of $p_{crossover}=0.5)$. Then, the offspring are subjected to a polynomial mutation process (following the same probability distribution from the SBX with $p_{mutation}=0.05$). With this, we get a new generation of individuals that will be evaluated, repeating this process of selection, crossover, and mutation until it reaches $G=700$ generations. 
    At the end of the evolutionary optimisation, the best individuals are selected for post-evaluation. These individuals are re-evaluated for $S$ trials, hence obtaining a better estimation of the average first passage time $t_f$ produced by the EBN, as done for the RBN experiments.

    \subsection{Measuring Chaos in BNs}
    \label{sec:delta}
    There are several ways to measure chaos in BNs~\cite{Carlos04Chaos, Zang2016}. In our case, we measure it by analysing the sensitivity of the network dynamics to the initial conditions, following a methodology presented in~\cite{Carlos04Chaos}. A small perturbation is introduced with a random initial state $A$, creating a new initial state $B$ by flipping one random node in the network. Two independent dynamics are then recorded starting from $A$ and $B$, and after $T$ time steps, two final states---$A'$ and $B'$---are obtained. To determine if the introduced perturbation affects the dynamics of the network, we first compute the normalised Hamming distance between the $[A,B]$ and $[A', B']$ as follows:
   \begin{equation}\label{eq:hamming}
        H(A,B)=\frac{1}{N}\sum_{i=1}^{N}|a_i-b_i|
    \end{equation}
    where $a_i$ ($b_i$) is the state of the $i^\text{th}$ node of network $A$ ($B$). Then, we compute the variation $\Delta$ introduced by the perturbation as follows:
        \begin{equation}\label{eq:delta}
        \Delta=H(A',B')-H(A,B) = H(A',B') - \frac{1}{N}
    \end{equation}    
    This way, we can understand whether the network dynamics amplify the perturbation or the states converge. A negative $\Delta$ means the Hamming distance was reduced, indicating the network tends to the same attractor, presenting an ordered regime. A positive $\Delta$ means the network diverges if initialized with different states, even when very similar, indicating that a small perturbation can affect its dynamics and, therefore, present an edge-of-chaos or chaotic regime.
    
    We create $100$ random initial and related perturbed states for each network. We record the final states after $T=10^4$ time steps and compute $\Delta$ according to \eqref{eq:delta}, obtaining the average from all the runs. By performing this analysis, we can infer if a BN presents a chaotic or edge-of-chaos regime, and we can correlate this finding to the network's performance to understand whether dynamic could be beneficial for exploration in our setting.

\section{Results and Discussion}
\label{Results}
    
\begin{figure}[h!]
        \centering
        \includegraphics[width=3.4in]{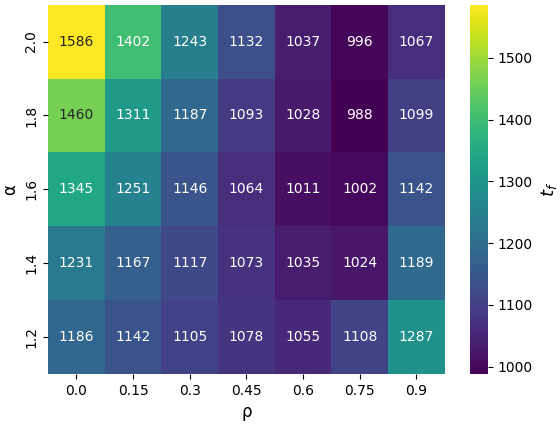}
        \caption{Performance in terms of average first passage time $t_f$ obtained with the LMCRW approach systematically varying the value of the parameters $\rho$ and $\alpha$. The reported values correspond to the mean over 20 independent evaluations.}
        \label{fig:crwlevyHeatmap}
    \end{figure}

    First, we compute the performance of the baseline approach, where a swarm of kilobots executes the LMCRW with fixed parameters $\rho$ and $\alpha$ governing, respectively, the turning angle and the step length distribution. We perform an extensive analysis to determine the best combination of these parameters, varying $\rho\in\{0,0.15,0.3,0.45,0.6,0.75,0.9\}$ and $\alpha\in\{1.2,1.4,1.6,1.8,2.0\}$. For each combination, we ran $20$ evaluations, trying to minimize any possible bias. As said in Section \ref{experimental_setup}, an evaluation corresponds to $S=100$ simulation trials with different target positions and robots' initial positions. In Figure~\ref{fig:crwlevyHeatmap}, the average value over the 20 evaluations is reported, indicating that the lowest first passage time $t_f$ is obtained by setting $\rho=0.75$ and $\alpha=1.8$, substantially confirming the results obtained in \cite{Dimidov2016}. This configuration is selected for comparison with the RBN and EBN approaches discussed below. 

    \subsection{Random Boolean Network}

\begin{figure}[h]
        \centering
        \includegraphics[width=3.4in]{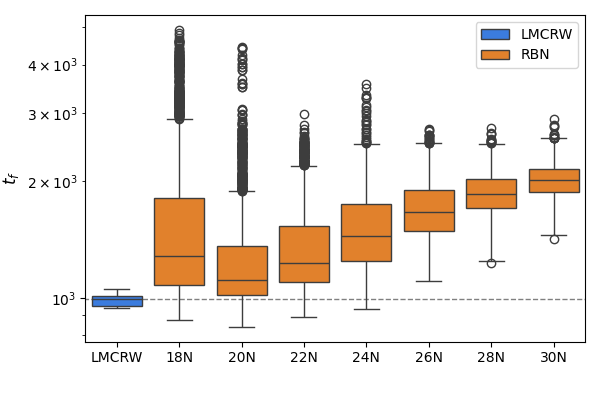}
        \caption{Comparison between LMCRW method with the RBN in terms of the average first passage time $t_f$, computed over 20 independent evaluations. For the RBNs, all the 100 created networks are grouped on their respective boxplots. The y axis is in log scale and the dashed line represents the LMCRW median.}
        \label{fig:comparingAll}
    \end{figure}

    We evaluate how efficient RBNs drive the Kilobots' exploration by evaluating different network sizes with $N\in[18, 20, 22, 24, 26, 28,30]$ nodes. We generate $100$ different RBN for each size, and each RBN is tested over 20 independent evaluations, the same process used for evaluating the LMCRW approach. In Figure~\ref{fig:comparingAll}, we grouped all the 2000 data points obtained in each boxplot. First and foremost, we note that the performance of the RBNs changes with the network size $N$ and is highly variable depending on the RBN configuration. Looking at $t_f$ for the $2000$ values presented for each RBN size, we can see that the 30N RBN presents the worst performance with an average $t_f$ of $\unit[2.016\times10^3]{s}$ and standard deviation of $\unit[0.209\times10^3]{s}$. In contrast, the best results are presented by the 20 nodes network (20N), with an average $t_f$ of $\unit[1.3\times10^3]{s}$ and standard deviation of $\unit[0.492\times10^3]{s}$.

    
    \begin{table}[h]
        \centering
        \caption{Performance comparison between RBNs and LMCRW. According to the statistical test (Bonferroni p-value = 0.0005), we group RBNs for having statistically significant differences (labeled ``worse'' or ``better'') or not (labeled ``similar').}
        \label{tab:RBNperf}
            \begin{tabularx}{3.4in}{X c c c}
                \hline
                Nodes $N$ & Worse & Similar & Better \\
                \hline
                $N=18$ & 87\% & 12\% & 1\%  \\
                $N=20$ & 65\% & 33\% & 2\% \\
                $N=22$ & 91\% & 8\% & 1\% \\
                $N=24$ & 98\% & 2\% & 0\% \\
                $N=26$ & 100\% & 0\% & 0\% \\
                $N=28$ & 100\% & 0\% & 0\% \\
                $N=30$ & 100\% & 0\% & 0\% \\ 
                \hline
            \end{tabularx}
    \end{table}
 
    To evaluate the efficiency of each RBN, we compare their performance with the baseline LMCRW approach. When both distributions are Normal, we use the T-test comparison. Otherwise, we use the Mann-Whitney U test. For both cases, we use a confidence interval of $95\%$, relying on the Bonferroni correction to deal with the Family Wise Error Rate (FWER) caused by multiple comparisons, and a Power of Test \cite{myors2010statistical} of $95\%$ for differences greater than $1.2$ standard deviation, as we have $N=20$ samples. In this way, we can measure the statistical difference between approaches, obtaining a more reliable percentage of networks that are better, similar (no statistical difference), or worse than the solutions found with the LMCRW approach (see Table~\ref{tab:RBNperf}). This comparison is made for each $100$ RBNs of each network size tested.
    
    This analysis reveals that for $N\in\{18,20,22\}$, RBNs with better performance than the baseline LMCRW approach were generated. No statistical difference was found for some RBNs up to $N=24$, while larger configurations generated only networks worse than the baseline.  
    The lower performance displayed by larger networks can be justified considering that the network size also determines the range of the step lengths that can be produced, as the maximum step-length size for a network is $2^{\frac{N}{2}}$ (see also Section~\ref{experimental_setup}). The larger the step length, the more probable it is to find robots colliding against the walls. For instance, with $N=30$ the maximum straight motion duration corresponds to  $2^{15}$ robot ticks. Considering that $\unit[1]{s}$ corresponds to 32 robot ticks and that a Kilobot moves with a velocity of about $\unit[1]{cm/s}$, the maximum step length is about $\unit[1024]{cm}$, while the experimental arena has a radius of just $r=\unit[45]{cm}$. Additionally, disentangling from collisions with walls or other robots would take longer, as this can happen only upon a sufficiently large rotation after the straight movement is terminated.  

    \begin{figure*}[t]
        \centering
        \includegraphics[width=\textwidth]{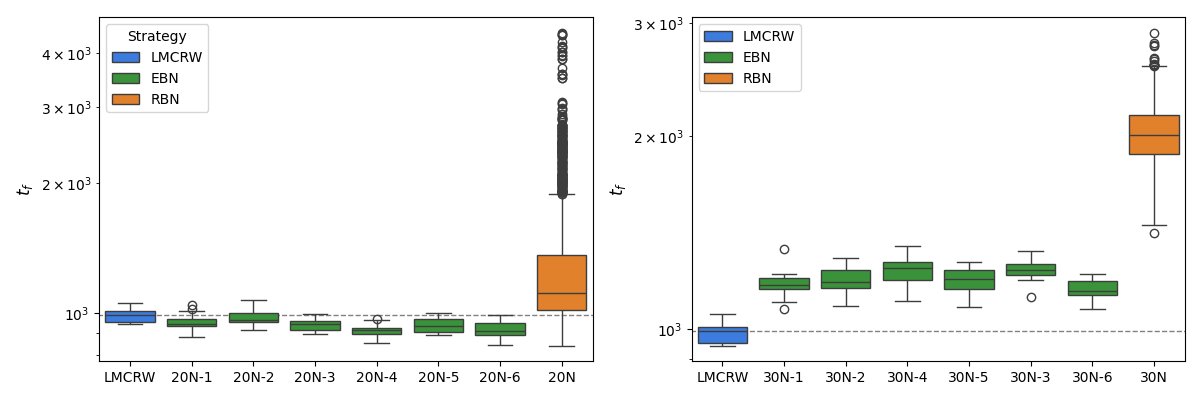}
        \caption{Comparison between RBNs and all 6 EBNs runs for $N=20,30$ with the LMCRW method. The y axis is in log scale and the dashed line represents the LMCRW median.}
        \label{fig:comparingRBNnEBN}
    \end{figure*}
    
    \subsection{Evolved Boolean Network}
    As mentioned before, while some RBNs can achieve a better performance than the baseline LMCRW approach, we notice a very high variability in the exploration performance of all networks. An evolutionary optimisation approach can help to find the best BN configuration systematically. In total, we performed six evolutionary runs for each network size. The number of runs was defined considering the high computational cost and time required for running each experiment replication. Then, the EBN with the best average $t_f$ after the post-evaluation results was selected for each run.

    
    \begin{table}[!b]
        \centering
        \caption{Performance comparison between EBNs and LMCRW. According to the statistical test (Bonferroni p-value = 0.0083), we group EBNs for having statistically significant differences (labeled ``worse'' or ``better'') or not (labeled ``similar').}
        \label{tab:EBNperf}
            \begin{tabularx}{3.4in}{X c c c}
                \hline
                Nodes $N$ & Worse & Similar & Better \\
                \hline
                $N=18$ & 0\% & 50\% & 50\%  \\
                $N=20$ & 0\% & 16.7\% & 83.3\%  \\
                $N=22$ & 33.3\% & 33.3\% & 33.3\% \\
                $N=24$ & 33.3\% & 16.7\% & 50.0\% \\
                $N=26$ & 33.3\% & 33.3\% & 33.3\%  \\
                $N=28$ & 100\% & 0\% & 0\% \\
                $N=30$ & 100\% & 0\% & 0\% \\
                \hline
            \end{tabularx}        
    \end{table}

    In Table~\ref{tab:EBNperf}, we report the percentage of EBNs that resulted in worse, similar, or better performance compared to the baseline LMCRW approach, using the same approach as with RBNs above, with a 95\% confidence interval and applying Bonferroni correction for the number of comparisons. The best results are obtained with $N=20$: evolution produces 4 out of 6 ($66.6\%$) networks with better first passage time $t_f$ than the baseline. The performances of the other two networks were not statistically different, indicating they are similar to the baseline. The second best size was $N=18$, which produces 3 out of 6 ($50\%$) better networks than LMCRW method, and 3 similar networks. EBNs with size $N=22$ and $N=26$ had a similar performance, producing 2 out of 6 ($33.3\%$) better, similar, and worse solutions than LMCRW. The evolution for EBN with $N=24$ produced 3 out of 6 ($50\%$) better solutions. Despite this setup producing one more better solution than $N=22$ and $N=24$, the relation between the network size and the capacity of the evolutionary process to produce good solutions seems to be the most important factor in the overall proportion of better, similar, and worse generated solutions. Surprisingly, the GA could improve the $N=24$ and $N=26$, even if its RBN counterpart could not produce worse results than the baseline. In contrast, when $N=28$ and $N=30$, all EBNs performed worse than the baseline LMCRW approach.
    The best $N=20$ network run achieves an average first passage time $t_f$ of $\unit[0.913\times10^3]{s}$ with a standard deviation of $\unit[28.06]{s}$, which represents $7.6\%$ improvement when comparing with the average $\unit[0.988\times10^3]{s}$ of LMCRW (which has $\unit[31.25]{s}$ of standard deviation). 
    
    We can analyse the impact of the evolution for $N=20,30$ by looking at Figure~\ref{fig:comparingRBNnEBN}. It is possible to notice that the evolutionary optimisation for $N=20$ led to much less variability in the results. Still, no run could significantly improve the performance compared with the best results of the RBN with the same size. On the other hand, for $N=30$, the GA reduced the variability and improved the performance by a great margin for all evolutionary runs. Despite that improvement, the results for this network size are still worse than the baseline.

    \subsection{Behaviour Analisys}

    \begin{figure*}[ht!]
        \centering
        \includegraphics[width=\textwidth]{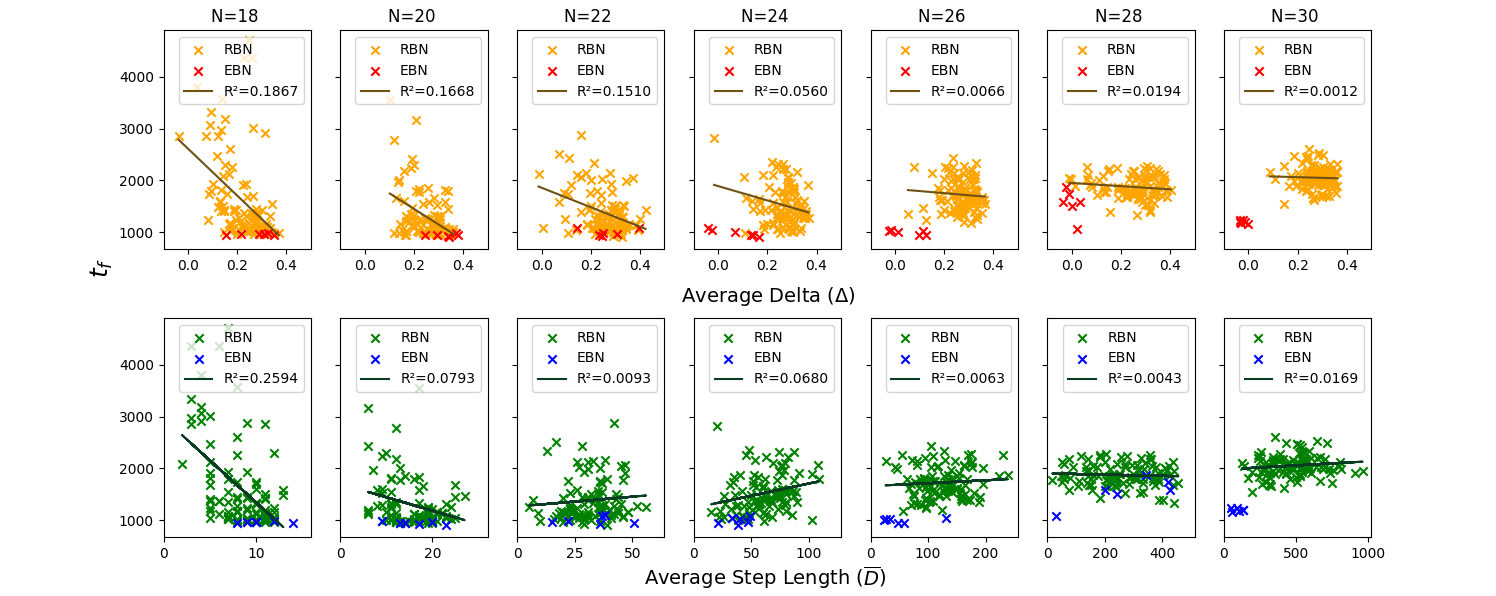}
        \caption{Correlation between performance ($t_f$) and sensitivity to initial conditions ($\Delta$, top panels) and average straight motion duration ($\overline{D}$, bottom panels) for RBNs and EBNs with $N\in\{18, 20, 22, 24, 26, 30\}$.}
        \label{fig:fptxchaos}
    \end{figure*}
    
     One of the hypotheses for the good performance of BNs compared to the baseline LMCRW approach is that BNs can produce chaotic or edge-of-chaos dynamics. Hence, seeking to measure the relation between performance and chaos, we compute for all the RBNs the correlation between the sensitivity to perturbations $\Delta$ defined in Section~\ref{sec:delta} and the average first passage time $t_f$ (see the top panels in Figure~\ref{fig:fptxchaos}) using a simple linear regression. Additionally, we also consider the correlation between the average straight motion duration, $\overline{D}$, generated by each network during exploration, and the average first passage time $t_f$ (see the bottom panels in Figure~\ref{fig:fptxchaos}). 
     
    \begin{figure}[h!]
        \centering
        \includegraphics[width=3.4in]{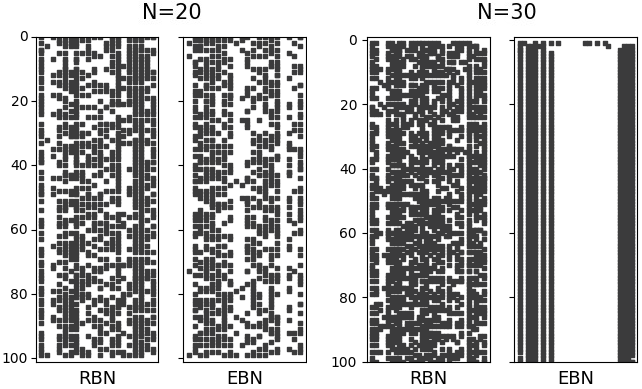}
        \caption{Node values (horizontal dimension) across time (vertical dimension)for EBN and RBN of 20N and 30N. White represents 0, and gray represents 1. For each network size, the RBN corresponds to the best network out of the 100 networks shown in Figure~\ref{fig:comparingAll}, while the EBN is the best network over the six evolutionary runs. }
        \label{fig:attractorsGraph}
    \end{figure}
    
     The RBNs with $N=18$ have the highest negative correlation between $\Delta$ and $t_f$, indicating that effective exploration (i.e., low $t_f$) corresponds to chaotic dynamics of the network (i.e., high $\Delta$). The correlation begins to fade when the network size grows and, consequently, the maximum straight motion duration. This indicates that chaotic behaviour has more influence when the RBN does not produce straight movements that could exceed the arena size and frequently expose robots to collisions. For the EBNs, we notice that for $N=18,20,22$ the behaviour tends to be chaotic ($\Delta$ approximately between 0.2 and 0.4). Still, with the growth of the network size, evolution produces more ordered networks (low $\Delta$ values). Interestingly, we can see a similar pattern in the correlation between $\overline{D}$ and $t_f$. It is easy to notice with these figures that for larger networks, the GA produces a more ordered behaviour to keep the $\overline{D}$ small, reducing the collisions within the walls and so improving the performance for the corresponding EBN. For network sizes of $N=20$ and $N=22$, evolution does not need to constrain the $\overline{D}$, so it has more freedom to search for more effective behavior patterns, exploiting the chaotic properties of the networks (i.e., higher $\Delta$ values).



    
    
    To better understand the effect of the evolutionary process on the BN's chaotic properties, we looked at the activation patterns over time for networks of size $N=20$ and $N=30$ both without and with evolution (see Figure~\ref{fig:attractorsGraph}). 
    For $N=30$, the EBN converges to an ordered regime of one attractor with 1 state. Interestingly, the average step length presented in this attractor is $58 cm$, which is very close to the size of the arena radius. As previously presented, we can assume that the evolutionary process simplifies the behaviour of the network to a more ordered regime to keep the step-length sizes small. By doing this, robots do not make long steps and avoid getting stuck against the wall, resulting in better performance. The EBN with $N=20$ performs way better since this network size produces step lengths that are fitter to the specific arena size used in this task. We can see in Figure~\ref{fig:attractorsGraph} that this EBN presents an edge-of-chaos/chaotic regime. This is aligned with the results shown in Figure~\ref{fig:fptxchaos}: there is a positive correlation between chaos and performance when the step length is appropriate with respect to the size of the environment to be explored. Since this network has a maximum step length value of $32 cm$, the evolutionary process can freely explore the solutions based on the edge-of-chaos dynamics for improving the network, selecting the ones that present more chaotic properties (i.e., higher $\Delta$ values).

 \section{Conclusion}\label{Conclusions}




Tasks in which a given area has to be explored are important for a great number of applications, including agriculture, search and rescue, and surveillance among others. The use of robots, for that end, has the potential to make exploration faster and safer, mainly in environments that are obnoxious to humans. Among the different possible strategies, swarm robotics offers the possibility of using a swarm of minimalist, therefore inexpensive and replaceable, robots for a fast, effective, and robust exploration. To achieve that, the main challenge is to define the proper behavioral strategies for every single individual that will form the swarm behavior.

Due to the natural similarity with many biological individuals that effectively behave in a swarm, such as ants, bees, birds, and so on, bio-inspired strategies are usually employed in swarm robotics. In this work, we investigate whether the use of very commonly found dynamics in exploratory processes in nature, i.e., chaotic and edge-of-chaos dynamics, can improve the exploratory capacity of a swarm of simple robots. We use RBNs, which is a well-known model in the literature capable of producing this kind of dynamics. By comparing these dynamics with LMCRW, a traditional and widely-used method for exploration in swarm robotics, we found evidence that indeed confirms our initial hypothesis that chaotic and edge-of-chaos dynamics would lead to faster exploration. Although the RBNs can outperform the LMCRW approaches, this is very dependent on a good random initialization of the network, which leads to highly variable performance results on multiple runs. In order to overcome this limitation, we resort to an evolutionary algorithm for defining the network connections and boolean functions. Our results show that evolved BNs produce improved exploratory abilities. Such improvement is dependent on the relation between the sizes of the network and the area to be explored. The evolutionary process can also remove the dependency on a good random initialization for all boolean network sizes, guaranteeing the best performance level of each size when compared with their RBN counterparts. Furthermore, the best network size leads to significantly shorter exploration with respect to the best LMCRW approach. Notably, the BN critical regime was maintained. We also demonstrated that in this scenario, there is a positive correlation between the network sensitivity to perturbations---a property strongly linked to critical dynamics---and the swarm exploratory performance.

Overall, these results highlight an interesting research direction regarding the use of chaotic dynamics in swarm robotics exploratory tasks, especially when using minimalistic robots. Future work includes testing the best strategies using real robots, developing more efficient algorithms to optimize the boolean networks, and studying more complex scenarios in which robots have minimal collective perception capabilities.

\IEEEpeerreviewmaketitle

\bibliographystyle{IEEEtran}
\bibliography{references}

\begin{thebibliography}{10}
\providecommand{\url}[1]{#1}
\csname url@samestyle\endcsname
\providecommand{\newblock}{\relax}
\providecommand{\bibinfo}[2]{#2}
\providecommand{\BIBentrySTDinterwordspacing}{\spaceskip=0pt\relax}
\providecommand{\BIBentryALTinterwordstretchfactor}{4}
\providecommand{\BIBentryALTinterwordspacing}{\spaceskip=\fontdimen2\font plus
\BIBentryALTinterwordstretchfactor\fontdimen3\font minus
  \fontdimen4\font\relax}
\providecommand{\BIBforeignlanguage}[2]{{%
\expandafter\ifx\csname l@#1\endcsname\relax
\typeout{** WARNING: IEEEtran.bst: No hyphenation pattern has been}%
\typeout{** loaded for the language `#1'. Using the pattern for}%
\typeout{** the default language instead.}%
\else
\language=\csname l@#1\endcsname
\fi
#2}}
\providecommand{\BIBdecl}{\relax}
\BIBdecl

\bibitem{garate:2021}
B.~Garate, S.~D{\'\i}az, S.~Iturriaga, S.~Nesmachnow, V.~Shepelev, and
  A.~Tchernykh, ``Autonomous swarm of low-cost commercial unmanned aerial
  vehicles for surveillance,'' \emph{Programming and Computer Software},
  vol.~47, no.~8, pp. 558--577, 2021.

\bibitem{polvara:21}
R.~Polvara, F.~Del~Duchetto, G.~Neumann, and M.~Hanheide, ``Navigate-and-seek:
  a robotics framework for people localization in agricultural environments,''
  \emph{IEEE Robotics and Automation Letters}, vol.~6, no.~4, pp. 6577--6584,
  2021.

\bibitem{10.1109/jproc.2021.3072740}
M.~Dorigo, G.~Theraulaz, and V.~Trianni, ``{Swarm Robotics: Past, Present, and
  Future},'' \emph{Proceedings of the IEEE}, vol. 109, no.~7, pp. 1152--1165,
  2021.

\bibitem{10.1126/scirobotics.abe4385}
------, ``{Reflections on the future of swarm robotics},'' \emph{Science
  Robotics}, vol.~5, no.~49, p. eabe4385, 2020.

\bibitem{7067029}
M.~Sitti, H.~Ceylan, W.~Hu, J.~Giltinan, M.~Turan, S.~Yim, and E.~Diller,
  ``Biomedical applications of untethered mobile milli/microrobots,''
  \emph{Proceedings of the IEEE}, vol. 103, no.~2, pp. 205--224, 2015.

\bibitem{10.1038/s41578-018-0001-3}
M.~Sitti, ``{Miniature soft robots — road to the clinic},'' \emph{Nature
  Reviews Materials}, vol.~3, no.~6, pp. 74--75, 2018.

\bibitem{10.1002/adma.201703554}
H.~Zeng, P.~Wasylczyk, D.~S. Wiersma, and A.~Priimagi, ``Light robots: Bridging
  the gap between microrobotics and photomechanics in soft materials,''
  \emph{Advanced Materials}, vol.~30, no.~24, p. 1703554, 2018.

\bibitem{Gauci:2014kb}
M.~Gauci, J.~Chen, W.~Li, T.~J. Dodd, and R.~Gross, ``{Self-organized
  aggregation without computation},'' \emph{The International Journal of
  Robotics Research}, vol.~33, no.~8, pp. 1145 -- 1161, 2014.

\bibitem{8264725}
A.~Özdemir, M.~Gauci, S.~Bonnet, and R.~Groß, ``Finding consensus without
  computation,'' \emph{IEEE Robotics and Automation Letters}, vol.~3, no.~3,
  pp. 1346--1353, 2018.

\bibitem{10.1109/lra.2022.3150479}
L.~Feola and V.~Trianni, ``{Adaptive Strategies for Team Formation in
  Minimalist Robot Swarms},'' \emph{IEEE Robotics and Automation Letters},
  vol.~7, no.~2, pp. 4079--4085, 2021.

\bibitem{10.1126/scirobotics.aaw9710}
K.~N. McGuire, C.~{de Wagter}, K.~Tuyls, H.~J. Kappen, and G.~C. H.~E. {de
  Croon}, ``{Minimal navigation solution for a swarm of tiny flying robots to
  explore an unknown environment},'' \emph{Science Robotics}, vol.~4, no.~35,
  p. eaaw9710, 2019.

\bibitem{10.1126/scirobotics.abd8668}
F.~Berlinger, M.~Gauci, and R.~Nagpal, ``{Implicit coordination for 3D
  underwater collective behaviors in a fish-inspired robot swarm},''
  \emph{Science Robotics}, vol.~6, no.~50, p. eabd8668, 2021.

\bibitem{10.3389/frobt.2021.618268}
M.~Kegeleirs, G.~Grisetti, and M.~Birattari, ``{Swarm SLAM: Challenges and
  Perspectives},'' \emph{Frontiers in Robotics and AI}, vol.~8, p. 618268,
  2021.

\bibitem{bartumeus2005animal}
F.~Bartumeus, M.~G.~E. da~Luz, G.~M. Viswanathan, and J.~Catalan, ``Animal
  search strategies: a quantitative random-walk analysis,'' \emph{Ecology},
  vol.~86, no.~11, pp. 3078--3087, 2005.

\bibitem{Codling2008}
E.~Codling, M.~Plank, and S.~Benhamou, ``Random walks in biology,''
  \emph{Journal of the Royal Society, Interface / the Royal Society}, vol.~5,
  pp. 813--34, 09 2008.

\bibitem{Levy2015}
\BIBentryALTinterwordspacing
V.~Zaburdaev, S.~Denisov, and J.~Klafter, ``L\'evy walks,'' \emph{Rev. Mod.
  Phys.}, vol.~87, pp. 483--530, Jun 2015. [Online]. Available:
  \url{https://link.aps.org/doi/10.1103/RevModPhys.87.483}
\BIBentrySTDinterwordspacing

\bibitem{Angelani_2010}
L.~Angelani and R.~D. Leonardo, ``Geometrically biased random walks in
  bacteria-driven micro-shuttles,'' \emph{New Journal of Physics}, vol.~12,
  no.~11, p. 113017, 2010.

\bibitem{Dimidov2016}
C.~Dimidov, G.~Oriolo, and V.~Trianni, ``Random walks in swarm robotics: An
  experiment with kilobots,'' in \emph{Swarm Intelligence}, M.~Dorigo,
  M.~Birattari, X.~Li, M.~L{\'o}pez-Ib{\'a}{\~{n}}ez, K.~Ohkura, C.~Pinciroli,
  and T.~St{\"u}tzle, Eds.\hskip 1em plus 0.5em minus 0.4em\relax Cham:
  Springer International Publishing, 2016, pp. 185--196.

\bibitem{Hecker2015}
J.~Hecker and M.~Moses, ``Beyond pheromones: evolving error-tolerant, flexible,
  and scalable ant-inspired robot swarms,'' \emph{Swarm Intelligence}, vol.~9,
  03 2015.

\bibitem{Kegeleirs2019}
M.~Kegeleirs, D.~Garz{\'o}n~Ramos, and M.~Birattari, ``Random walk exploration
  for swarm mapping,'' in \emph{Towards Autonomous Robotic Systems},
  K.~Althoefer, J.~Konstantinova, and K.~Zhang, Eds.\hskip 1em plus 0.5em minus
  0.4em\relax Cham: Springer International Publishing, 2019, pp. 211--222.

\bibitem{Reina2015}
A.~Reina, R.~Miletitch, M.~Dorigo, and V.~Trianni, ``A quantitative micro-macro
  link for collective decision: the shortest path discovery/selection
  example,'' \emph{Swarm Intelligence}, 05 2015.

\bibitem{10.48550/arxiv.2305.16063}
M.~Raoufi, P.~Romanczuk, and H.~Hamann, ``{Individuality in Swarm Robots with
  the Case Study of Kilobots: Noise, Bug, or Feature?}'' in \emph{ALIFE 2023:
  Ghost in the Machine: Proceedings of the 2023 Artificial Life
  Conference}.\hskip 1em plus 0.5em minus 0.4em\relax MIT Press, 2023, p.
  isal\_a\_00623.

\bibitem{kilobots}
M.~Rubenstein, C.~Ahler, N.~Hoff, A.~Cabrera, and R.~Nagpal, ``Kilobot: A low
  cost robot with scalable operations designed for collective behaviors,''
  \emph{Robotics and Autonomous Systems}, vol.~62, no.~7, pp. 966 -- 975, 2014,
  reconfigurable Modular Robotics.

\bibitem{Pinciroli:2018}
C.~Pinciroli, M.~S. Talamali, A.~Reina, J.~A.~R. Marshall, and V.~Trianni,
  ``{Simulating Kilobots Within ARGoS: Models and Experimental Validation},''
  in \emph{Swarm Intelligence}, ser. Lecture Notes in Computer Sciences,
  M.~Dorigo, M.~Birattari, C.~Blum, A.~L. Christensen, A.~Reina, and
  V.~Trianni, Eds., vol. 11172.\hskip 1em plus 0.5em minus 0.4em\relax Cham:
  Springer International Publishing, 2018, pp. 176--187.

\bibitem{kaplan1997}
D.~Kaplan and L.~Glass, \emph{Understanding Nonlinear Dynamics}, ser. Textbooks
  in Mathematical Sciences.\hskip 1em plus 0.5em minus 0.4em\relax Springer New
  York, 1997.

\bibitem{CarlosRBN}
C.~Gershenson, ``Introduction to random boolean networks,'' 08 2004.

\bibitem{LANGTON1990}
C.~G. Langton, ``Computation at the edge of chaos: Phase transitions and
  emergent computation,'' \emph{Physica D: Nonlinear Phenomena}, vol.~42,
  no.~1, pp. 12--37, 1990.

\bibitem{west:03}
M.~J. West-Eberhard, \emph{Developmental plasticity and evolution}.\hskip 1em
  plus 0.5em minus 0.4em\relax Oxford University Press, 2003.

\bibitem{morales:21}
G.~B. Morales and M.~A. Mu{\~n}oz, ``Optimal input representation in neural
  systems at the edge of chaos,'' \emph{Biology}, vol.~10, no.~8, p. 702, 2021.

\bibitem{clark:20}
T.~Clark and A.~D. Luis, ``Nonlinear population dynamics are ubiquitous in
  animals,'' \emph{Nature ecology \& evolution}, vol.~4, no.~1, pp. 75--81,
  2020.

\bibitem{zhang:20}
X.~Zhang, Z.~Wu, and L.~Chua, ``Hearts are poised near the edge of chaos,''
  \emph{International Journal of Bifurcation and Chaos}, vol.~30, no.~09, p.
  2030023, 2020.

\bibitem{ruan:19}
H.~Ruan, Q.~Sun, W.~Zhang, Y.~Liu, and L.~Lai, ``Targeting intrinsically
  disordered proteins at the edge of chaos,'' \emph{Drug discovery today},
  vol.~24, no.~1, pp. 217--227, 2019.

\bibitem{VOLOS2013}
C.~Volos, I.~Kyprianidis, and I.~Stouboulos, ``Experimental investigation on
  coverage performance of a chaotic autonomous mobile robot,'' \emph{Robotics
  and Autonomous Systems}, vol.~61, no.~12, pp. 1314--1322, 2013.

\bibitem{DaRold2015}
F.~Da~Rold, ``Deterministic chaos in mobile robots,'' 07 2015, pp. 1--7.

\bibitem{Pinciroli:2012dc}
C.~Pinciroli, V.~Trianni, R.~O'Grady, G.~Pini, A.~Brutschy, M.~Brambilla,
  N.~Mathews, E.~Ferrante, G.~D. Caro, F.~Ducatelle, M.~Birattari, L.~M.
  Gambardella, and M.~Dorigo, ``{ARGoS: a modular, parallel, multi-engine
  simulator for multi-robot systems},'' \emph{Swarm Intelligence}, vol.~6,
  no.~4, pp. 271 -- 295, 2012.

\bibitem{kaplan1958nonparametric}
E.~L. Kaplan and P.~Meier, ``Nonparametric estimation from incomplete
  observations,'' \emph{Journal of the American statistical association},
  vol.~53, no. 282, pp. 457--481, 1958.

\bibitem{KAUFFMAN1969}
S.~Kauffman, ``Metabolic stability and epigenesis in randomly constructed
  genetic nets,'' \emph{Journal of Theoretical Biology}, vol.~22, no.~3, pp.
  437--467, 1969.

\bibitem{Shmulevich2009}
I.~Shmulevich and E.~R. Dougherty, \emph{Probabilistic Boolean Networks: The
  Modeling and Control of Gene Regulatory Networks}.\hskip 1em plus 0.5em minus
  0.4em\relax USA: Society for Industrial and Applied Mathematics, 2009.

\bibitem{Derrida1986}
B.~Derrida and Y.~Pomeau, ``Random networks of automata: A simple annealed
  approximation,'' \emph{Europhysics Letters ({EPL})}, vol.~1, no.~2, pp.
  45--49, jan 1986.

\bibitem{Roli2011}
A.~Roli, M.~Manfroni, C.~Pinciroli, and M.~Birattari, ``On the design of
  boolean network robots,'' vol. 6624, 04 2011, pp. 43--52.

\bibitem{Garattoni2013}
L.~Garattoni, A.~Roli, M.~Amaducci, C.~Pinciroli, and M.~Birattari, ``Boolean
  network robotics as an intermediate step in the synthesis of finite state
  machines for robot control,'' vol.~12, 09 2013, pp. 783--790.

\bibitem{Roli2013}
A.~Roli, M.~Villani, R.~Serra, L.~Garattoni, C.~Pinciroli, and M.~Birattari,
  ``Identification of dynamical structures in artificial brains: An analysis of
  boolean network controlled robots,'' vol. 8249, 12 2013, pp. 324--335.

\bibitem{mitchell1998}
M.~Mitchell, \emph{An introduction to genetic algorithms}.\hskip 1em plus 0.5em
  minus 0.4em\relax MIT press, 1998.

\bibitem{Ching2008AGA}
\BIBentryALTinterwordspacing
W.-K. Ching, H.~Leung, S.~Zhang, and N.-K. Tsing, ``A genetic algorithm for
  optimal control of probabilistic boolean networks,'' 2008. [Online].
  Available: \url{https://api.semanticscholar.org/CorpusID:11624190}
\BIBentrySTDinterwordspacing

\bibitem{roli2011boolean}
A.~Roli, C.~Arcaroli, M.~Lazzarini, and S.~Benedettini, ``Boolean networks
  design by genetic algorithms,'' 2011.

\bibitem{GADeb2007}
\BIBentryALTinterwordspacing
K.~Deb, K.~Sindhya, and T.~Okabe, ``Self-adaptive simulated binary crossover
  for real-parameter optimization,'' ser. GECCO '07.\hskip 1em plus 0.5em minus
  0.4em\relax New York, NY, USA: Association for Computing Machinery, 2007, p.
  1187–1194. [Online]. Available:
  \url{https://doi.org/10.1145/1276958.1277190}
\BIBentrySTDinterwordspacing

\bibitem{Carlos04Chaos}
C.~Gershenson, C.~L. Apostel, B.~K. B, and B.~Belgium, ``Phase transitions in
  random boolean networks with different updating schemes,'' 2004.

\bibitem{Zang2016}
X.~Zang, S.~Iqbal, Y.~Zhu, X.~Liu, and J.~Zhao, ``Applications of chaotic
  dynamics in robotics,'' \emph{International Journal of Advanced Robotic
  Systems}, vol.~13, no.~2, p.~60, 2016.

\bibitem{myors2010statistical}
B.~Myors, K.~R. Murphy, and A.~Wolach, \emph{Statistical power analysis: A
  simple and general model for traditional and modern hypothesis tests}.\hskip
  1em plus 0.5em minus 0.4em\relax Routledge, 2010.

\end{thebibliography}



%








\end{document}